\documentclass[letterpaper, 10 pt, conference]{ieeeconf}  % Comment this line out if you need a4paper

\IEEEoverridecommandlockouts                              % This command is only needed if 
\usepackage{graphics} % for pdf, bitmapped graphics files
\usepackage{epsfig} % for postscript graphics files
\usepackage{mathptmx} % assumes new font selection scheme installed
\usepackage{times} % assumes new font selection scheme installed
\usepackage{amsmath} % assumes amsmath package installed
\usepackage{amssymb}  % assumes amsmath package installed
\usepackage{algorithm}  
\usepackage{algorithmic}
\usepackage{booktabs}
\usepackage{multirow}
\usepackage{subfigure}
\usepackage{xcolor}

\usepackage[normalem]{ulem}
\useunder{\uline}{\ul}{}

\newcommand{\Eqref}[1]{Eq.(\ref{#1})}
\newcommand*\rot{\rotatebox{90}}

\title{\LARGE \bf
Few-shot 3D LiDAR Semantic Segmentation for Autonomous Driving
}

\author{Jilin Mei$^{1}$, Junbao Zhou$^{1,2}$ and Yu Hu$^{1,2,*}$% <-this % stops a space
\thanks{\dag This work was supported by National Natural Science Foundation of China under Grant No.62203424 and No.62176250.}% <-this % stops a space
\thanks{$^{1}$Research Center for Intelligent Computing Systems, Institute of Computing Technology, Chinese Academy of Sciences, Beijing, 100190, China. }%
\thanks{$^{2}$School of Computer Science and Technology, University of Chinese Academy of Sciences, Beijing, 100049, China. }%
\thanks{$^{*}$Correspondence: Yu Hu, huyu@ict.ac.cn}%
}

\begin{document}

\maketitle
\thispagestyle{empty}
\pagestyle{empty}

%%%%%%%%%%%%%%%%%%%%%%%%%%%%%%%%%%%%%%%%%%%%%%%%%%%%%%%%%%%%%%%%%%%%%%%%%%%%%%%%
\begin{abstract}
In autonomous driving, the novel objects and lack of annotations challenge the traditional 3D LiDAR semantic segmentation based on deep learning. Few-shot learning is a feasible way to solve these issues. However, currently few-shot semantic segmentation methods focus on camera data, and most of them only predict the novel classes without considering the base classes. This setting cannot be directly applied to autonomous driving due to safety concerns. Thus, we propose a few-shot 3D LiDAR semantic segmentation method that predicts both novel and base classes simultaneously. Our method tries to solve the background ambiguity problem in generalized few-shot semantic segmentation. We first review the original cross-entropy and knowledge distillation losses,  then propose a new loss function that incorporates the background information to achieve 3D LiDAR few-shot semantic segmentation. Extensive experiments on SemanticKITTI demonstrate the effectiveness of our method.
\end{abstract}

%%%%%%%%%%%%%%%%%%%%%%%%%%%%%%%%%%%%%%%%%%%%%%%%%%%%%%%%%%%%%%%%%%%%%%%%%%%%%%%%
\section{INTRODUCTION}
3D LiDAR can provide accurate depth information, making it an important sensor for autonomous driving systems \cite{LiMZLCCL21}. Recently, researchers have focused on semantic segmentation of LiDAR point clouds \cite{chen2021rangeseg, wang2020speed, Chen2022RangeSegRR}, which primarily involves deep learning with large-scale annotation data \cite{Gao2022AreWH}. As autonomous driving scenarios are dynamic and complex, deep semantic segmentation faces the following challenges: 1) emergence of novel objects; traditional methods require training before testing; as a result, only classes defined in the training stage can be processed, and it is difficult to identify novel objects. 2) lack of annotations; pixel-level annotation requires a lot of resources, and it is difficult to collect large-scale annotation samples. To meet these challenges, few-shot semantic segmentation plays a significant role in the perception of autonomous driving.

Currently, most research on few-shot semantic segmentation \cite{Xie2021FewShotSS, DING2022109018, Liu2022AxialAC} focuses on image datasets, such as PASCAL VOC \cite{PSACALVOC} and MS COCO \cite{COCO14}. As part of this task, the data is divided into base and novel classes. During training, the base classes have many samples, while the novel class has few (e.g., 1-10) samples. During testing, the classical few-shot semantic segmentation only recognizes novel classes and ignores the base classes; obviously, this setting can not be applied directly to autonomous driving. For example, if the novel class is cyclists, recognizing only this class cannot ensure driving safety. As a result, generalized few-shot semantic segmentation methods are proposed, that is, to recognize both the base  and novel classes simultaneously during the testing stage. Generalized methods \cite{tian2022generalized,MyersDean2021GeneralizedFS}, including meta-learning and transfer learning, are discussed gradually for image data.

In generalized few-shot semantic segmentation, one problem is the background ambiguity: the background data of the current training step may contain the foreground objects of the other steps and vice versa, which decreases the model's accuracy and generalization. Point cloud datasets collected from LiDAR, like SemanticKITTI \cite{SemanticKITTI}, usually have contains a variety of objects in a single frame, and the background ambiguity needs to be considered. Therefore, the key to solving few-shot LiDAR semantic segmentation is how to model background information. According to our literature review, it is found that few works have addressed this specific issue, especially those based on LiDAR data.

This paper assumes that semantic information about the background changes continuously with the training process, that is, a pixel/point may belongs to the background in the current training step, but belongs to the foreground in the next. The paper first reviews the original cross-entropy and knowledge distillation losses, then proposes a new loss function that takes into account the background information to achieve LiDAR few-shot semantic segmentation. We have carried out extensive experiments on SemanticKITTI \cite{SemanticKITTI} dataset, and the proposed method significantly outperforms the ordinary transfer learning based method. In summary, the contributions of this paper include:
       \begin{itemize}
            \item To the best of our knowledge, we are the first to discuss the background ambiguity for few-shot 3D LiDAR semantic segmentation.
            \item We introduce the unbiased cross-entropy loss and knowledge distillation loss to explicitly tackle the background ambiguity, which improves the model's accuracy and generalization.
            \item Experiments show that our method outperforms
baseline methods and is effective for few-shot 3D LiDAR semantic segmentation, with a noticeable improvement for novel classes. 
        \end{itemize}

\section{RELATED WORK}
\subsection{Few-shot Semantic Segmentation for Image Data}
Semantic segmantation for image data is a widely researched problem, while deep neural network is the dominant solution for semantic segmentation \cite{long2015fully} , there are still a few works based on traditional machine learning algorithms \cite{wang2020residual, wang2020g, wang2019wavelet}.

Few-shot learning for semantic segmentation was introduced in 2017. Shaban et al. \cite{shaban2017one} propose few-shot semantic segmentation based on deep learning and design a two-branch network. The first branch converts the support images and ground truths into prototype vectors; the second branch inputs the prototype vectors and the query images and outputs the prediction of masks. This two-branch design has been widely adopted in subsequent works and has developed into a classic few-shot semantic segmentation method \cite{Rakelly2018Conditional, ZhangLLYS19, SiamOJ19}. Wang et al. \cite{WangLZZF19}  improve the model generalization by aligning prototype vectors with alternate support sets and query sets. Wang et al. \cite{WangY0ZS021} utilize the probabilistic implicit variable model to model the probability distribution of the prototype. Yang et al. \cite{YangLLJY20} uses multiple prototype features to correspond to different image regions, thus improving their semantic representation ability. Zhang et al. \cite{ZhangXQ21} decouple the prototype features into main feature vectors and auxiliary feature vectors to improve the segmentation accuracy. In terms of problem definition, a lot of methods only predict the mask of novel classes. However, autonomous driving systems need to effectively identify all objects in the scene. Obviously, the classical few-shot semantic segmentation can not directly meet this requirement.

Recently, researchers have begun paying attention to the generalized few-shot semantic segmentation method. The generalized method predicts the masks for both base and novel classes simultaneously. A variety of technical routes have been presented in relevant research. Yan et al. \cite{YanCXWLL19} extend Faster/Mask R-CNN by proposing meta-learning over RoI features instead of a full image feature. Tian et al. \cite{tian2022generalized} propose the first prototype-based learning method, which uses context information to build prototype features. Myers-Dean et al. \cite{MyersDean2021GeneralizedFS} explore the fine-tuning method, learning the initial weight using the support set, and then fine-tuning the query set. In addition, some researchers \cite{KhandelwalGS21, TaveraCMC22} discuss the domain-adaption method to solve the problem of lack of labeled samples in the target domain. Currently, most of the research on few-shot semantic segmentation focuses on image datasets such as Pascal VOC and MS COCO, while little attention is paid to LiDAR point cloud data used in autonomous driving.

\subsection{Few-shot Semantic Segmentation for Point Cloud Data}
Recently, the semantic segmentation based on LiDAR point clouds has drawn researchers' attention \cite{Gao2022AreWH}. In contrast, few-shot semantic segmentation based on LiDAR point clouds has not been extensively discussed. On the one hand, the dataset supporting this task is much smaller than the image data; on the other hand, LiDAR point clouds are sparse and unevenly distributed, making it difficult to apply existing image-based methods directly.

Zhao et al. \cite{ZhaoFewCVPR2021} incorporate EdgeConv and self-attention into the embedding network, and a k-NN graph is built to match query features and multi-prototype vectors. Lai et al. \cite{Lai2022TacklingBA} point out that in the process of parameter learning, background information is ambiguous, and the background point of one episode may be the foreground of another. Thus, using cross-entropy directly as a loss function will result in a degradation of accuracy. Finally,  \cite{Lai2022TacklingBA} adopts the entropy of predictions on query samples to the loss function as an additional regularization. Chen et al. \cite{Chen2020CompositionalPN} sample multiple two-dimensional perspectives of three-dimensional data and establish prototype features for each perspective. The above works are evaluated on dense and uniform point cloud data, and only the prediction of novel classes is considered during testing. Directly migrating these methods to sparse LiDAR data is difficult. Corral-Soto et al. \cite{CorralSoto2022HYLDAEH} tried to realize the few-shot semantic segmentation of LiDAR data using domain adaptation, which integrates self-supervised, unsupervised, and semi-supervised learning. Although it completes the domain adaptation task, it still needs hundreds of labeled samples, which is far more than the setting of few-shot learning.

\section{PROPOSED METHODOLOGY}
\subsection{Problem Definition}
For few-shot learning setting, let $C_{b}$ denote the set of base classes that contain a large number of data for each class, and $C_{n}$ be the set of novel classes which contains few data for each class. Additionally, the background/unknown class $u$ is excluded from $C_{b}$ and $C_{n}$. Thus, the overall class set $C=\left \{ u \right \}  \cup C_{b} \cup C_{n}$ and $C_{b}\cap C_{n}=\emptyset $. The training stage involves base data $D_{b}:\left \{ x_{b}^{i}, y_{b}^{i} \right \}_{i=1}^{P}$ and novel data $D_{n}:\left \{ x_{n}^{i}, y_{n}^{i} \right \}_{i=1}^{Q}$, where $y_{b}^{i} \mapsto  R^{M\times (\left | C_{b} \right |+1) } $ is the ground truth of $x_{b}^{i}$ and $y_{n}^{i} \mapsto R^{M\times (\left | C_{n} \right |+1) } $ is the ground truth of $x_{n}^{i}$, $M$ is the number of elements of $x^i$. In classical few-shot semantic segmentation, only the classes in $C_{n}$ are predicted during the testing stage, and the data belonging to $C_{b}$ is treated as background or unknown. As for generalized few-shot semantic segmentation, both $C_{n}$ and $C_{b}$ should be processed simultaneously during the testing stage. Autonomous driving system relies on LiDAR point clouds to capture the surrounding environment, so identifying only novel objects is insufficient to ensure safety. Thus, the problem is defined as a generalized few-shot semantic segmentation of LiDAR point clouds. 

\begin{figure*}
\centering
\includegraphics[width=0.8\textwidth]{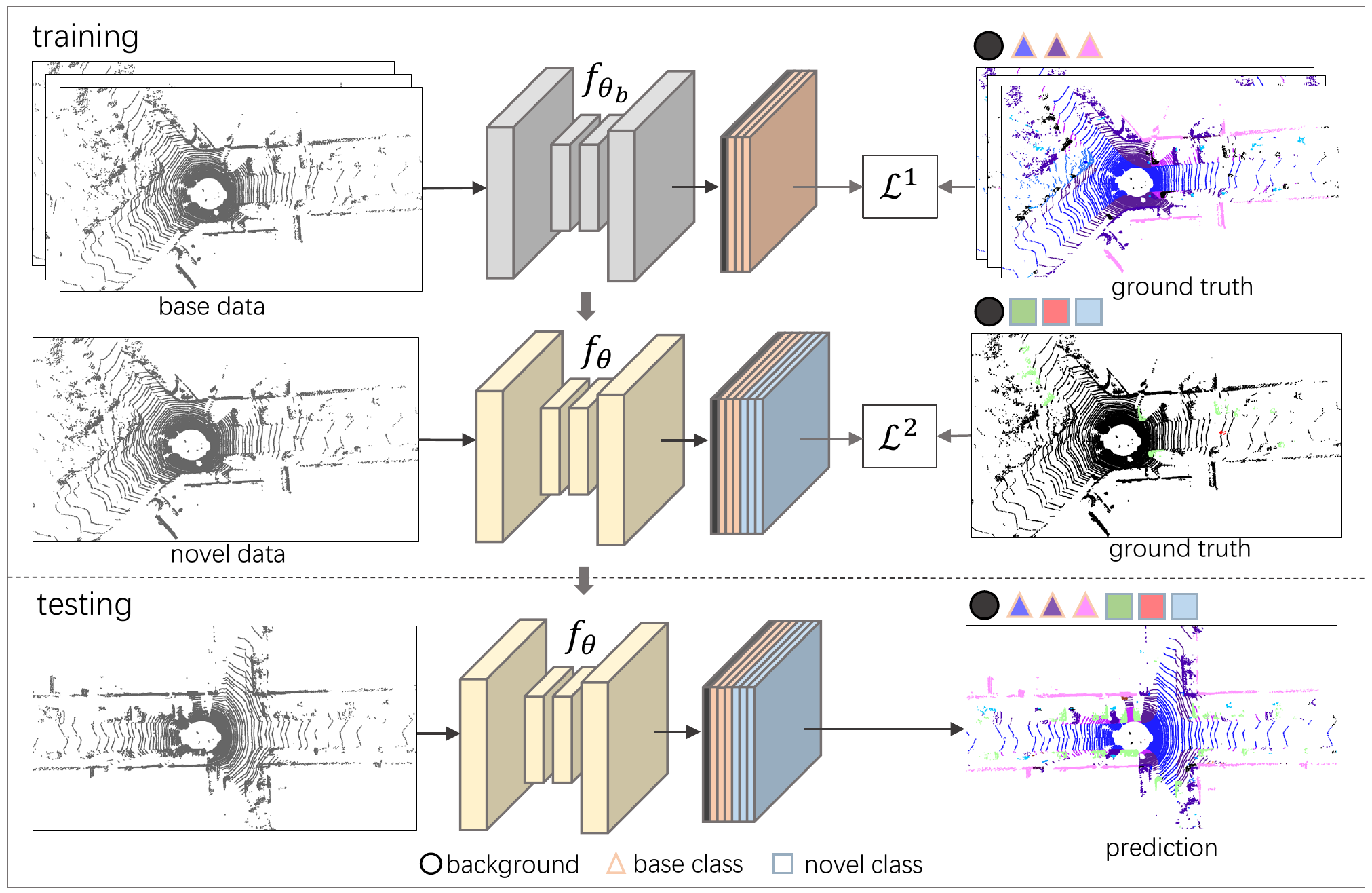}
\caption{\textbf{Overview of our proposed method.} In the training step, the background of base and novel data has different semantics, for example, the road(blue color) in the ground truth of base data is the foreground, but the road in novel data is defined as the background; the transfer learning is adopted for parameter updating during training, where the base and novel data share the same backbone network; to address the problem of background ambiguity, we design the loss function $L^1$ for base data and $L^2$ for novel data. In the testing step, the model predicts both novel and base classes simultaneously.}
\label{fig:network architecture}
\end{figure*}

As shown in Fig. \ref{fig:network architecture}, the problem of few-shot semantic segmentation is addressed with three steps based on transfer learning. In the first step, we update the model parameter $\theta_{b}$ for abundant base data:
\begin{align}
\hat{\theta}_{b}=\underset {\theta_{b}}{\operatorname{arg\,min}}\, L^1(X_b,Y_b;\theta_{b}^0),
\end{align}
where $\theta_{b}^0$ is the initial state of the model parameter, and $L^1$ is the loss function. After obtaining optimal $\theta_{b}$, the model parameter is adjusted with a few novel data in the second step:
\begin{align}
\hat{\theta}=\underset {\theta}{\operatorname{arg\,min}}\, L^2(X_n,Y_n;\theta_{b},\theta^0),
\end{align}
where $\theta^0$ is initialized with $\theta_{b}$, and $L^2$ is the loss function that explicitly considers the background ambiguity. Then the final model parameter $\theta$ is used for testing in the third step. The prediction of each input data $x^i$ is defined as:
\begin{align}
f_{\theta} \left ( x^i \right )\to  y^{i},
\end{align}
where $f$ denotes the model and $y^i \mapsto R^{M\times \left | C \right | }$.

\subsection{Base Model Training}
Let $C_b^{'}=\left \{ u \right \} \cup C_b$ be the classes in base model training, where $u$ is background. There are abundant data for each class in $C_b$, so supervised learning is adopted to update the network parameter $\theta_b$. 

Data imbalance directly affects the accuracy of the model. In the SemanticKITTI dataset, for example, the number of points on the road is 1000 times higher than that of person. When the data imbalance is not taken into account, the model's output will be biased towards the category that appears most frequently in training. For this reason, we use weighted cross-entropy loss as part of the loss function. For each LiDAR data frame, the loss item $L_{ce}^{1}$ is defined as:
\begin{align}
\label{eq-4}
L_{ce}^{1}=-\frac{1}{M}\sum_{i=1}^{M}\sum_{k\in C_{b}^{'}}^{}\alpha_k \textbf{1}  [ y_{b}^{i}(k)=1] \ln ( P_{\theta_b}^{k}(x_b^i)),      
\end{align}
where $M$ is the number of points, and the one-hot vector $y_{b}^{i}$ is the truth of point $x_b^i$. $P_{\theta_b}^{k}(x_b^i)$ is the probability of point $x_b^i$ belonging to class $k$, and let $P_{\theta_b}^{k}(x_b^i)$ be the output of model. $\alpha_k$ is the weight factor for class $k$, calculated as the inverse of the frequency.

During training, each pixel/point will be treated as independent of each other if the cross-entropy loss is used to adjust the parameters. However, the intersection over Union (IoU) measure is used to evaluate the model performance in testing. The inconsistency between the training and testing measures will lead to the decline of the model performance. Inspired by \cite{SalsaNext}, we introduce the $Lov\acute{a}sz \ Softmax$ loss term into the loss function to maximize the IoU score. Nevertheless, IoU is a discrete 
 measure that cannot be directly applied as a loss due to its inability to be derived. The authors of \cite{BermanTB18} use the $Lov\acute{a}sz$ extension for submodular functions to adopt this measure, and the final $Lov\acute{a}sz \ Softmax$ loss($L_{ls}$) is defined as:
\begin{align}
L_{ls}^1  =& \frac{1}{|C_{b}^{'}|}\sum_{k\in C_{b}^{'}}^{}\overline{\bigtriangleup_{J_k}}(m(k)), \\
m_i(k) =& \begin{cases}
 1-P_{\theta_b}^{k}(x_b^i) \ &if \  k  = y_b^i(k)\\ P_{\theta_b}^{k}(x_b^i) \  &otherwise
\end{cases}   
\end{align}
where $J_k$ defines the Jaccard index, and $\overline{\bigtriangleup_{J_k}}$ is $Lov\acute{a} sz$ extension of
the Jaccard index. 

The final loss function for base model training is defined as :
\begin{align}
L^{1}=L_{ce}^{1}(X_b,Y_b;\theta_{b})+L_{ls}^{1}(X_b,Y_b;\theta_{b}).
\end{align}

\subsection{Novel Data fine-tuning}
In order to incorporate novel classes, we use the transfer learning method as shown in Fig.\ref{fig:network architecture}. Our method uses a deep CNN model as the backbone network, and transforms the semantic segmentation task into a pixel-level classification process. Therefore, each class has its own classification header. As part of the fine-tuning process, the initial network parameters are copied directly from the base model. Additionally, $|C_n|$ classification headers are added in order to correspond with the novel classes. As shown in Fig.\ref{fig:network architecture}, the base classes are treated as background in the ground truth of novel data. In addition to following the classical few-shot learning model, this settings also meet the actual use case. Firstly, some cases may make it impossible to obtain the base data due to privacy concerns. Secondly, incremental parameter learning can be achieved if one only needs to label a few novel data to accurately recognize novel classes in autonomous driving systems, which will greatly improve the updating efficiency of the model.

In fact, the novel objects may exist in the background of the base data, and the base objects may also exist in the background of the novel data. Thus, if the background ambiguity is not considered, the model will be biased towards the novel classes after fine-tuning, which will result in a reduction in the overall accuracy of the model. Inspired by \cite{MiBCVPR2020}, we explicitly model the background information by modifying the loss function. This allows the background to be incorporated into both cross-entropy loss and knowledge distillation loss.

\subsubsection{Modeling Background in Cross-entropy Loss}
The catastrophic forgetting problem will occur in the base classes when the standard cross-entropy loss provided by \Eqref{eq-4} is directly used in the fine-tuning stage due to the base objects appearing in the background of the novel data. Therefore, we rewrite the \Eqref{eq-4} with base model parameter $\theta_b$:
\begin{align}
L_{ce}^{2}  = &-\frac{1}{M}\sum_{i=1}^{M} \sum_{k\in C_{n}^{'}}^{}\alpha_k\textbf{1}[ y_{n}^{i}(k)=1 ] \ln ( \hat{P}_{\theta}^{k}(x_n^i)  ) , \\
&\hat{P}_{\theta}^{k}(x_n^i)  = \begin{cases}
 P_{\theta}^{k}(x_n^i) \  &if \  k  \ne  u\\ \sum_{k^{'}\in C_b}^{} P_{\theta_b}^{k^{'}}(x_n^i) \ &if \  k=u
\label{eq-9}
\end{cases}   
\end{align}
where $C_{n}^{'}=\left \{ u \right \} \cup C_n$. Compared with \Eqref{eq-4}, we extent the output of model with $\hat{P}_{\theta}^{k}(x_n^i)$ in \Eqref{eq-9}. The main idea is to deal with the background and foreground separately, rather than combining them. For the foreground points in novel data, we directly use the prediction results of the updated model parameter $\theta$; for the background points in novel data, the base model parameter $\theta_b$ is used for prediction. Note that the base model here can only output the predictions for the base classes, then the predictions of all base classes are added and compared with the background ground truth of novel data.

\subsubsection{Modeling Background in Distillation Loss}
In traditional knowledge distillation(KD), the softened class scores of the teacher are used as the extra supervisory signal: the distillation loss encourages the student to mimic the scores of the teacher\cite{TungM19}. Compared to the one-hot label, the output from the teacher network contains more information about the fine-grained structure of data. Generally, the teacher network is larger than the student network, and both networks have the same dimensions of output. In the setting of our method, the teacher is the base model $f_{\theta_b}$ and the student is the fine-tuned model $f_{\theta}$, they share the same backbone structure but have a different number of classification heads. The purpose of introducing KD loss is to mitigate catastrophic forgetting during the fine-tuning stage of novel data. 
\begin{align}
\label{eq-10}
L_{kd}  = &-\frac{1}{M}\sum_{i=1}^{M} \sum_{k\in C_{b}^{'}}^{}P_{\theta_b}^k(x_n^i) \ln( \tilde{P}_{\theta}^{k}(x_n^i)  ) 
\end{align}

\Eqref{eq-10} shows the KD loss, the one-hot ground truth is replaced with the output of the base model. There are two potential issues with directly using the original KD loss: 1) as shown in Fig. \ref{fig:network architecture}, the output dimensions of $f_{\theta_b}$ and $f_{\theta}$ are inconsistent; 2) the background prediction in $f_{\theta_b}$ and $f_{\theta}$ has different semantics, for example, $f_{\theta}$ perceives the road as background, while $f_{\theta_b}$ perceives it as background in Fig. \ref{fig:network architecture}. Considering the above issues, the original KD loss can not adequately transfer the knowledge from $f_{\theta_b}$ to $f_{\theta}$. To this extent we define the distillation loss of novel data by rewriting $\tilde{P}_{\theta}^{k}(x_n^i)$ in \Eqref{eq-10} as: 
\begin{align}
\tilde{P}_{\theta}^{k}(x_n^i)  = \begin{cases}
 P_{\theta}^{k}(x_n^i), \  &if \  k  \ne  u\\ \sum_{k^{'}\in C_n}^{} P_{\theta}^{k^{'}}(x_n^i), \ &if \  k=u.
\end{cases}  
\label{eq-11}
\end{align}
Similarly to \Eqref{eq-9}, we explicitly consider the background in novel data. For the foreground($k \ne u$) points in novel data, we directly compare the output of $f_{\theta_b}$ and $f_{\theta}$; as for the background($k=u$) points, the outputs of $f_{\theta}$ on all novel classes are firstly accumulated, then compared with the output of $f_{\theta_b}$. 
\begin{align}
L_{ls}^2  = & \frac{1}{|C_{n}^{'}|}\sum_{k\in C_{n}^{'}}^{}\overline{\bigtriangleup_{J_k}}(m(k)), \\
m_i(k) = &  \begin{cases}
 1-P_{\theta}^{k}(x_n^i), & \  if \  k  = y_n^i(k)\\
 P_{\theta}^{k}(x_n^i), & \  otherwise.
\end{cases}   
\end{align}

Besides the loss items $L_{ce}^2$ and $L_{kd}$, the  $Lov\acute{a} sz \ Softmax$ loss $L_{ls}^2$ is also included during fine-tuning. Then the final loss function of $L^2$ is defined as :
\begin{align}
L^{2} & = L_{ce}^{2}(X_n,Y_n;\theta)+L_{ls}^{2}(X_n,Y_n;\theta)+L_{kd}(X_n,Y_n;\theta,\theta_b).
\end{align}

\subsection{Implement Details}
\label{III-D}
%Model
The backbone network is based on SalsaNext \cite{SalsaNext}, the proposed method focuses on designing the loss function to mitigate the problem of background ambiguity, thus our method can easily be integrated into other backbone networks. The raw LiDAR point clouds are projected to range view images to obtain a grid and dense data structure. For each LiDAR point $(x,y,z)$, the corresponding point $(u,v)$ in range image is defined as:
\begin{align}
\begin{pmatrix}
 u\\v
\end{pmatrix}  = \begin{pmatrix}
 \frac{1}{2}[1- \arctan(y,x)\pi^{-1} ]w \\ [1-(\arcsin(z, r^{-1})+f_{down})f^{-1}]h
\end{pmatrix}
\end{align}
where $w$ and $h$ denote the width and height of range image; $r=\sqrt{x^2+y^2+z^2}$, and $f=|f_{down}|+|f_{up}|$ defines the vertical view of LiDAR. The input of network is a tensor with the dimension $[w,h,5]$, containing the point $(x,y,z)$, the intensity value $\mathfrak{i}$ and range value $r$.

\section{Experiments}

\subsection{Experiment Settings}
%数据集和评价指标
%类别定义，base类有哪些，novel类有哪些
%实验配置，机器、训练的超参数
%各种baseline的解释

We evaluate the performance of our proposed method on SemanticKITTI \cite{SemanticKITTI} dataset, which is a large-scale LiDAR dataset providing over 43K annotated full 3D LiDAR scans. As described in Sect. \ref{III-D}, the backbone network is derived from SalsaNext \cite{SalsaNext}, but the uncertainty part is not included. The input dimension of network is $64\times2048\times5$.

We first divide the dataset into training and validation splits. Sequences between 00 and 10 are used for training and sequence 08 is used for validation. The SemanticKITTI includes 20 classes. We set the \textbf{car, person, bicyclist, and motorcyclist} as the novel classes and the other 16 classes as base classes. This setting mainly considers that dynamic objects usually appear in the novel class, which requires special attention in autonomous driving applications.  As discussed above, we adopt transfer learning for parameter updating in model training which contains two steps: base model training and novel data fine-tuning. In the base model training, we use the whole training split while the novel classes (car, person, bicyclist, and motorcyclist) are defined as background. In the novel data fine-tuning, considering the setting of few-shot learning, we do not make use of the whole training split, but randomly sample a certain number of scans from the training split, where the novel classes are defined as foreground and base classes are defined as background.

To fully demonstrate the performance of our method, we sample different shots per novel class for evaluation. One shot corresponds to one frame of LiDAR scan. We sample frames in the following manner, firstly we pick all the frames which contain a novel class in the training split, then for each novel class, we randomly sample $n$ frames($n = 10, 20, 50, 100$). Besides, to make sure the sampled frames keep consistent in every training process, we manually fix the random seed and apply it in all the training processes.

All the compared methods and ours are trained in the same environment with the same hyperparameters. We train all the models on NVIDIA V100 GPU, and we employ the SGD optimizer with the initial learning rate of 0.01 which is decayed by 0.01 after every epoch.  The momentum of the SGD optimizer and the batch size is set to 0.9 and 14, respectively. 

The performance of our proposed method is measured by mean intersection-over-union (mIoU), which is given by \Eqref{eq:mIoU}, where $P_i$ is model's prediction of class $i$, $G_i$ is the annotation set of class $i$. Note that we have split the classes into a base subset and a novel subset, we need to evaluate our method by $mIoU$ over base classes and $mIoU$ over novel classes, which are denoted as $mIoU_b$ and $mIoU_n$, respectively.

\begin{align}
mIoU = \frac{1}{C} \sum_{i=1}^{C} \frac{|P_i \cap G_i|}{|P_i \cup G_i|} \times 100
\label{eq:mIoU}
\end{align}

% \begin{figure*}
% \centering
% \includegraphics[width=0.9\textwidth]{fig/fig2.pdf}
% \caption{\textbf{The comparisons on the valid set of SemanticKITTI.} Left: the mIoU of base classes $mIoU_b$ for different methods. Mid: the mIoU of novel classes $mIoU_n$ for different methods. Right: the $mIoU$ for different methods.}
% \label{fig:fig2}
% \end{figure*}

\begin{table}[]

\begin{center}
\caption{\textbf{Quantitative Evaluations on validation set of SemanticKITTI.} $mIoU_{n}$ for 4 novel classes, $mIoU_{b}$ for 16 base classes and $mIoU$ for all classes.}

\tabcolsep=0.4cm
\renewcommand\arraystretch{1.3}
\label{tab:mIoU_result}
\begin{tabular}{|l|l|l|l|l|}
\hline
                          & method                                    & $mIoU_{b}$           & $mIoU_{n}$           & $mIoU$                 \\ \hline
                          & Base model                                & 58.7                & -                    & -                \\ \hline
\multirow{4}{*}{shot=100} & GFSS \cite{MyersDean2021GeneralizedFS}     & {57.1}          & 20.9                & 49.5                \\ \cline{2-5} 
                          & GFSS-dyn \cite{MyersDean2021GeneralizedFS} & 54.1                & 48.6                & 53.0                \\ \cline{2-5} 
                          & LwF \cite{Li2018LearningWF}                & 55.4                & 48.3                & 53.9                \\ \cline{2-5} 
                          & \textbf{Ours}                                      & \textbf{56.6}       & {\textbf{50.9}} & {\textbf{55.5}} \\ \hline
\multirow{3}{*}{shot=50}  & GFSS-dyn \cite{MyersDean2021GeneralizedFS} & 53.8                & 38.2                & 50.5                \\ \cline{2-5} 
                          & LwF \cite{Li2018LearningWF}                & 51.6                & 44.0                & 50.0                \\ \cline{2-5} 
                          & \textbf{Ours}                                      & {\textbf{55.3}} & {\textbf{47.5}} & {\textbf{53.7}} \\ \hline
\multirow{3}{*}{shot=20}  & GFSS-dyn \cite{MyersDean2021GeneralizedFS} & 52.1                & 20.0                & 45.4                \\ \cline{2-5} 
                          & LwF \cite{Li2018LearningWF}                & 49.2                & 34.1                & 46.1                \\ \cline{2-5} 
                          & \textbf{Ours}                                      & {\textbf{55.0}} & {\textbf{34.4}} & {\textbf{50.7}} \\ \hline
\multirow{3}{*}{shot=10}  & GFSS-dyn \cite{MyersDean2021GeneralizedFS} & 54.9                & 23.5                & 48.4                \\ \cline{2-5} 
                          & LwF \cite{Li2018LearningWF}                & 53.3                & {28.4}          & 48.0                \\ \cline{2-5} 
                          & \textbf{Ours}                                      & {\textbf{55.9}} & \textbf{24.2}       & {\textbf{49.2}} \\ \hline
\end{tabular}

\end{center}
\end{table}

\begin{figure*}
\centering
\includegraphics[width=0.9\textwidth]{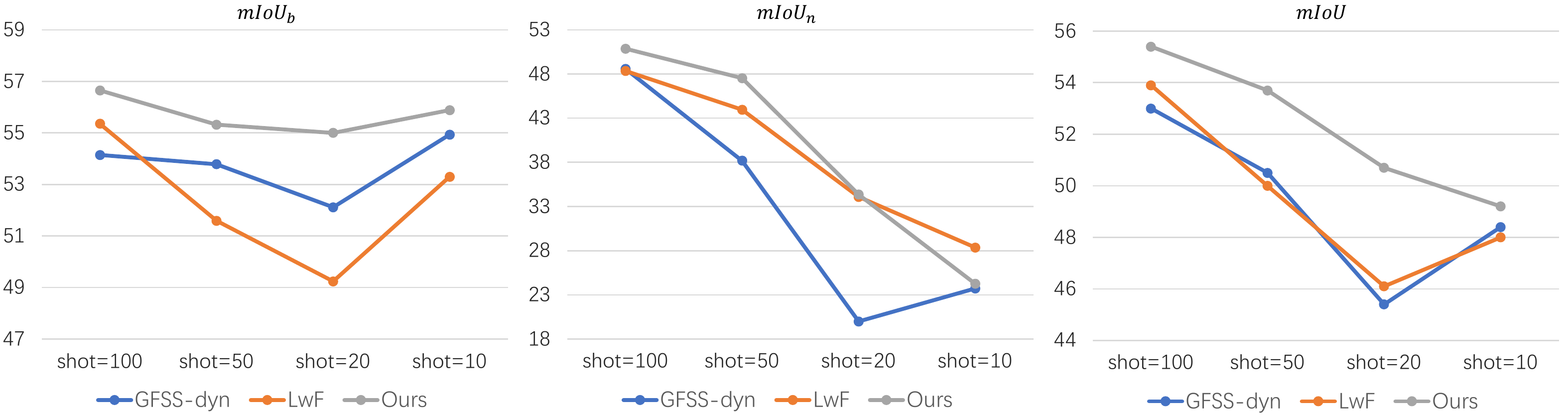}
\caption{\textbf{The comparisons on the validation set of SemanticKITTI.} Left: the mIoU of base classes $mIoU_b$ for different methods. Mid: the mIoU of novel classes $mIoU_n$ for different methods. Right: the $mIoU$ for different methods.}
\label{fig:line}
\end{figure*}

\begin{table*}[]

\tabcolsep=0.095cm
\renewcommand\arraystretch{1.2}
\caption{\textbf{Quantitative Evaluations on testing set of SemanticKITTI.} $mIoU_{n}$ for 4 novel classes, $mIoU_{b}$ for 16 base classes and $mIoU$ for all classes. }
\label{tab:all_test_iou}
\begin{tabular}{|l|llllllllllllllll|llll|lll|}
\hline
shot                 & \multicolumn{1}{l|}{method} & \multicolumn{1}{l|}{\rot{bicycle}} & \multicolumn{1}{l|}{\rot{motorcycle}} & \multicolumn{1}{l|}{\rot{truck}} & \multicolumn{1}{l|}{\rot{other-vehicle}} & \multicolumn{1}{l|}{\rot{road}} & \multicolumn{1}{l|}{\rot{parking}} & \multicolumn{1}{l|}{\rot{sidewalk}} & \multicolumn{1}{l|}{\rot{other-ground}} & \multicolumn{1}{l|}{\rot{building}} & \multicolumn{1}{l|}{\rot{fence}} & \multicolumn{1}{l|}{\rot{vegetation}} & \multicolumn{1}{l|}{\rot{trunk}} & \multicolumn{1}{l|}{\rot{terrain}} & \multicolumn{1}{l|}{\rot{pole}} & \rot{traffic-sign} & \multicolumn{1}{l|}{\rot{car}} & \multicolumn{1}{l|}{\rot{person}} & \multicolumn{1}{l|}{\rot{bicyclist}} & \rot{motorcyclist} & \multicolumn{1}{l|}{$mIoU_b$} & \multicolumn{1}{l|}{$mIoU_n$} & $mIoU$              \\ \hline
\multirow{4}{*}{100} & GFSS                        & 41.4                               & 28.1                                  & 27.2                             & 22.6                                     & 89.9                            & 57.5                               & 73.5                                & 27.3                                    & 85.1                                & 53.4                             & 77.8                                  & 61.3                             & 63.5                               & 49.9                            & 56.3               & 77.6                           & 4.9                               & 1.8                                  & 0.6                & {54.3}                    & 21.2                          & 47.4                \\
                     & GFSS-dyn                    & 41.2                               & 27.0                                  & 28.2                             & 15.8                                     & 88.7                            & 56.7                               & 70.4                                & 14.9                                    & 86.3                                & 52.4                             & 76.5                                  & 59.8                             & 59.5                               & 49.1                            & 54.6               & 81.5                           & 43.8                              & 40.5                                 & 23.3               & 52.1                          & 47.3                          & 50.1                \\
                     & LwF                         & 36.8                               & 28.1                                  & 23.4                             & 16.6                                     & 88.3                            & 55.2                               & 68.9                                & 19.6                                    & 86.8                                & 53.1                             & 77.6                                  & 60.7                             & 58.7                               & 48.8                            & 54.3               & 80.1                           & 42.3                              & 41.9                                 & 19.5               & 51.8                          & 46.0                          & 50.6                \\
                     & \textbf{Ours}                        & \textbf{34.0}                               & \textbf{30.2}                                  & \textbf{33.0}                             & \textbf{16.3}                                     & \textbf{89.7}                            & \textbf{56.9}                               & \textbf{71.7}                                & \textbf{17.9}                                    & \textbf{86.8}                                & \textbf{55.7}                             & \textbf{78.7}                                  & \textbf{60.9}                             & \textbf{63.6}                               & \textbf{52.8}                            & \textbf{54.2}               & \textbf{86.4}                           & \textbf{40.3}                              & \textbf{45.9}                                 & \textbf{17.6}               & \textbf{53.5}                 & {\textbf{47.6}}           & {\textbf{52.2}} \\ \hline
\multirow{3}{*}{50}  & GFSS-dyn                    & 37.2                               & 25.6                                  & 17.0                             & 13.8                                     & 88.1                            & 53.1                               & 69.0                                & 21.3                                    & 81.7                                & 49.4                             & 76.3                                  & 57.9                             & 61.8                               & 46.5                            & 51.0               & 75.3                           & 31.3                              & 22.0                                 & 12.6               & 50.0                          & 35.3                          & 46.9                \\
                     & LwF                         & 36.2                               & 23.2                                  & 19.2                             & 15.5                                     & 88.4                            & 54.0                               & 70.5                                & 10.3                                    & 83.5                                & 48.9                             & 76.9                                  & 57.3                             & 64.5                               & 48.5                            & 49.7               & 78.8                           & 35.5                              & 33.5                                 & 8.3                & 49.8                          & 39.0                          & 47.5                \\
                     & \textbf{Ours}                        & \textbf{29.8}                               & \textbf{28.1}                                  & \textbf{26.0}                             & \textbf{14.1}                                     & \textbf{89.4}                            & \textbf{59.7}                               & \textbf{70.6}                                & \textbf{22.1}                                    & \textbf{85.3}                                & \textbf{54.1}                             & \textbf{77.8}                                  & \textbf{59.7}                             & \textbf{62.1}                               & \textbf{52.7}                            & \textbf{54.8}               & \textbf{84.0}                           & \textbf{38.4}                              & \textbf{34.2}                                 & \textbf{15.9}               & {\textbf{52.4}}           & {\textbf{43.1}}           & {\textbf{50.5}} \\ \hline
\multirow{3}{*}{20}  & GFSS-dyn                    & 24.8                               & 16.5                                  & 20.1                             & 14.4                                     & 86.8                            & 50.1                               & 67.1                                & 20.3                                    & 79.0                                & 43.4                             & 73.8                                  & 55.9                             & 59.0                               & 40.5                            & 52.8               & 71.7                           & 0.1                               & 3.7                                  & 0.4                & 47.0                          & 19.0                          & 41.1                \\
                     & LwF                         & 23.1                               & 18.4                                  & 1.8                              & 18.4                                     & 87.6                            & 56.0                               & 68.0                                & 23.6                                    & 82.3                                & 47.1                             & 76.3                                  & 59.1                             & 59.0                               & 48.7                            & 53.6               & 74.6                           & 19.8                              & 20.9                                 & 7.5                & 48.2                          & 30.7                          & 44.5                \\
                         & \textbf{Ours}                        & \textbf{21.6}                               & \textbf{24.4}                                  & \textbf{24.5}                             & \textbf{19.5}                                     & \textbf{89.6}                            & \textbf{57.7}                               & \textbf{70.8}                                & \textbf{26.2}                                    & \textbf{84.9}                                & \textbf{52.6}                             & \textbf{78.6}                                  & \textbf{61.1}                             & \textbf{60.9}                               & \textbf{49.9}                            & \textbf{50.1}               & \textbf{81.8}                           & \textbf{20.5}                              & \textbf{16.1}                                 & \textbf{6.6}                & {\textbf{51.5}}           & {\textbf{31.3}}           & {\textbf{47.2}} \\ \hline
\multirow{3}{*}{10}  & GFSS-dyn                    & 25.3                               & 25.9                                  & 24.5                             & 18.4                                     & 89.1                            & 54.2                               & 71.2                                & 28.3                                    & 84.0                                & 49.7                             & 75.6                                  & 59.4                             & 61.9                               & 49.0                            & 55.6               & 73.6                           & 3.2                               & 8.7                                  & 0.0                & 51.5                          & 21.4                          & 45.1                \\
                     & LwF                         & 20.6                               & 23.3                                  & 32.8                             & 19.6                                     & 89.4                            & 54.6                               & 71.0                                & 24.9                                    & 85.6                                & 52.4                             & 77.2                                  & 59.8                             & 61.9                               & 45.6                            & 51.2               & 77.8                           & 15.2                              & 13.2                                 & 5.2                & 51.3                          & {27.9}                    & {46.4}          \\
                     & \textbf{Ours}                        & \textbf{22.1}                               & \textbf{29.2}                                  & \textbf{28.5}                             & \textbf{19.5}                                     & \textbf{89.4}                            & \textbf{56.2}                               & \textbf{69.9}                                & \textbf{27.0}                                    & \textbf{85.5}                                & \textbf{53.6}                             & \textbf{77.5}                                  & \textbf{60.6}                             & \textbf{60.5}                               & \textbf{50.4}                            & \textbf{53.5}               & \textbf{79.4}                           & \textbf{1.7}                               & \textbf{8.1}                                  & \textbf{0.8}                & {\textbf{52.2}}           & \textbf{22.5}                 & \textbf{46.0}       \\ \hline
\end{tabular}
\end{table*}

\begin{table}[]
\begin{center}
\caption{\textbf{Ablation study for different loss settings on testing set with shot=100.} $L_{ce}^{*}$ and $L_{kd}^{*}$ denote the original cross-entropy and knowledge distillation loss. $L_{ce}$ and $L_{kd}$ denote the proposed losses.}

\tabcolsep=0.3cm
\renewcommand\arraystretch{1.3}
\label{tab:ablation}

\begin{tabular}{llll|lll}
\hline
$L_{ce}^{*}$ & $L_{kd}^{*}$ & $L_{ce}$     & $L_{kd}$     & $mIoU_b$ & $mIoU_n$ & $mIoU$ \\ \hline
$\checkmark$ &              &              &              &   54.3  &  21.2   & 47.4  \\ \hline
$\checkmark$ & $\checkmark$ &              &              &   51.8  &  46.0   & 50.6  \\ \hline
$\checkmark$ &              &              & $\checkmark$ &   51.7  &  41.8   & 49.6  \\ \hline
             & $\checkmark$ & $\checkmark$ &              &   50.6  &  46.3   & 49.7  \\ \hline
             &              & $\checkmark$ &              &   52.7  &  46.1   & 51.4  \\ \hline
             &              & $\checkmark$ & $\checkmark$ &   53.5  &  47.6   & 52.2  \\ \hline
\end{tabular}
\end{center}
\end{table}

\subsection{Baselines}
We employ three methods as our baselines.
\begin{itemize}
    \item \textbf{GFSS} \cite{MyersDean2021GeneralizedFS}: it is a fine-tuning based method for generalized few-shot semantic segmentation. After base model training, it freezes the parameters in the backbone and only updates the parameters in classifiers with novel data. Our method shares the same training stage with GFSS, the improvement of ours is the consideration of background ambiguity in fine-tuning.
    \item \textbf{GFSS-dyn} \cite{MyersDean2021GeneralizedFS}: it has the same configurations as GFSS, except that the parameters of the backbone are dynamic during fine-tuning.
    \item \textbf{LwF} \cite{Li2018LearningWF}: it is a classic method in incremental learning, introducing knowledge distillation loss to prevent catastrophic forgetting. In our method, we also introduce knowledge distillation loss to achieve generalized few-shot semantic segmentation and prevent the mIoU of base classes does not largely decrease during fine-tuning.
\end{itemize}

\subsection{Quantitative Results}
%定量
We firstly analyze the quantitative comparisons on the validation set of SemanticKITTI and the results are reported in Table \ref{tab:mIoU_result} and Fig. \ref{fig:line}. At first, the original GFSS is implemented for LiDAR data with shot=100, but its performance at novel classes is only 20.9 which is largely lower than other methods; then we try to make the parameters of backbone network dynamic during novel data fine-tuning, and this strategy obtains a more reasonable result denoted as GFSS-dyn; thus, we only evaluate the GFSS-dyn for the other comparisons. The $mIoU_b$ of base model is 58.7. After novel data fine-tuning, our method performs better than others at $mIoU_b$, for example, the lowest scores of GFSS-dyn, LwF, and ours are 52.1, 49.2 and 55.0. For the measure of $mIoU_n$, it indicates the capability of learning new concepts when only a few samples are available. In Table \ref{tab:mIoU_result}, our method has higher $mIoU_n$ almost in all settings. Finally, the few-shot 3D LiDAR semantic segmentation should consider the base and novel classes at the same time, then the $mIoU$ over all classes is shown in Table \ref{tab:mIoU_result}. Our proposed method considerably outperforms the others by leading to the highest $mIoU$ score at all settings, especially, it is +1.5 over LwF and +2.4 over GFSS-dyn at shot=100. 

Table \ref{tab:all_test_iou} gives the IoU for each class on testing set. In most cases, our method has a higher mIoU than other methods.

\subsection{Ablation Study}
Our method explicitly incorporates the background information into the original cross-entropy loss and knowledge distillation loss. In Table \ref{tab:ablation}, the ablation study shows the effectiveness of our method.  When introducing $L_{ce}$, it improves the $mIoU$. And combining $L_{ce}$ and $L_{kd}$ brings the highest mIoU on testing set.

\section{CONCLUSIONS}
In this work, we formulate the few-shot 3D LiDAR semantic segmentation under the framework of transfer learning. With the help of knowledge distillation, our method can predict novel and base classes at the same time. To address the problem of background ambiguity, the proposed method explicitly incorporates the background information into the original cross-entropy loss and knowledge distillation loss. Despite its simplicity, our method achieves better performance on few-shot 3D LiDAR semantic segmentation settings, outperforming the previous transfer learning based methods. We hope this work will inspire further research on few-shot 3D LiDAR semantic segmentation.

%                                   % the textheight of the last page by a suitable amount.
%                                   % This command does not take effect until the next page
%                                   % so it should come on the page before the last. Make
%                                   % sure that you do not shorten the textheight too much.

\clearpage

\bibliographystyle{IEEEtran}
\bibliography{IEEEabrv,myref}

\begin{thebibliography}{10}
\providecommand{\url}[1]{#1}
\csname url@samestyle\endcsname
\providecommand{\newblock}{\relax}
\providecommand{\bibinfo}[2]{#2}
\providecommand{\BIBentrySTDinterwordspacing}{\spaceskip=0pt\relax}
\providecommand{\BIBentryALTinterwordstretchfactor}{4}
\providecommand{\BIBentryALTinterwordspacing}{\spaceskip=\fontdimen2\font plus
\BIBentryALTinterwordstretchfactor\fontdimen3\font minus
  \fontdimen4\font\relax}
\providecommand{\BIBforeignlanguage}[2]{{%
\expandafter\ifx\csname l@#1\endcsname\relax
\typeout{** WARNING: IEEEtran.bst: No hyphenation pattern has been}%
\typeout{** loaded for the language `#1'. Using the pattern for}%
\typeout{** the default language instead.}%
\else
\language=\csname l@#1\endcsname
\fi
#2}}
\providecommand{\BIBdecl}{\relax}
\BIBdecl

\bibitem{LiMZLCCL21}
Y.~Li, L.~Ma, Z.~Zhong, F.~Liu, M.~A. Chapman, D.~Cao, and J.~Li, ``Deep
  learning for lidar point clouds in autonomous driving: {A} review,''
  \emph{IEEE Transactions on Neural Networks and Learning Systems}, vol.~32,
  no.~8, pp. 3412--3432, 2021.

\bibitem{chen2021rangeseg}
T.-H. Chen and T.~S. Chang, ``Rangeseg: range-aware real time segmentation of
  3d lidar point clouds,'' \emph{IEEE Transactions on Intelligent Vehicles},
  vol.~7, no.~1, pp. 93--101, 2021.

\bibitem{wang2020speed}
G.~Wang, J.~Wu, R.~He, and B.~Tian, ``Speed and accuracy tradeoff for lidar
  data based road boundary detection,'' \emph{IEEE/CAA Journal of Automatica
  Sinica}, vol.~8, no.~6, pp. 1210--1220, 2020.

\bibitem{Chen2022RangeSegRR}
T.-H. Chen and T.~S. Chang, ``Rangeseg: Range-aware real time segmentation of
  3d lidar point clouds,'' \emph{IEEE Transactions on Intelligent Vehicles},
  vol.~7, pp. 93--101, 2022.

\bibitem{Gao2022AreWH}
B.~Gao, Y.~Pan, C.~Li, S.~Geng, and H.~Zhao, ``Are we hungry for 3d lidar data
  for semantic segmentation? {A} survey of datasets and methods,'' \emph{IEEE
  Transactions on Intelligent Transportation Systems}, vol.~23, no.~7, pp.
  6063--6081, 2022.

\bibitem{Xie2021FewShotSS}
G.~Xie, H.~Xiong, J.~Liu, Y.~Yao, L.~Shao, and M.~bin Zayed, ``Few-shot
  semantic segmentation with cyclic memory network,'' \emph{IEEE/CVF
  International Conference on Computer Vision}, pp. 7273--7282, 2021.

\bibitem{DING2022109018}
H.~Ding, H.~Zhang, and X.~Jiang, ``Self-regularized prototypical network for
  few-shot semantic segmentation,'' \emph{Pattern Recognition}, p. 109018,
  2022.

\bibitem{Liu2022AxialAC}
Y.~Liu, B.~Jiang, and J.~Xu, ``Axial assembled correspondence network for
  few-shot semantic segmentation,'' \emph{IEEE/CAA Journal of Automatica
  Sinica}, 2022.

\bibitem{PSACALVOC}
M.~Everingham, S.~M.~A. Eslami, L.~V. Gool, C.~K.~I. Williams, J.~M. Winn, and
  A.~Zisserman, ``The pascal visual object classes challenge: {A}
  retrospective,'' \emph{International Journal of Computer Vision}, vol. 111,
  no.~1, pp. 98--136, 2015.

\bibitem{COCO14}
T.~Lin, M.~Maire, S.~J. Belongie, J.~Hays, P.~Perona, D.~Ramanan,
  P.~Doll{\'{a}}r, and C.~L. Zitnick, ``Microsoft {COCO:} common objects in
  context,'' in \emph{European Conference on Computer Vision}, vol. 8693.\hskip
  1em plus 0.5em minus 0.4em\relax Springer, 2014, pp. 740--755.

\bibitem{tian2022generalized}
Z.~Tian, X.~Lai, L.~Jiang, S.~Liu, M.~Shu, H.~Zhao, and J.~Jia, ``Generalized
  few-shot semantic segmentation,'' in \emph{IEEE/CVF Conference on Computer
  Vision and Pattern Recognition}, 2022, pp. 11\,563--11\,572.

\bibitem{MyersDean2021GeneralizedFS}
J.~Myers-Dean, Y.~Zhao, B.~L. Price, S.~D. Cohen, and D.~Gurari, ``Generalized
  few-shot semantic segmentation: All you need is fine-tuning,'' \emph{ArXiv},
  vol. abs/2112.10982, 2021.

\bibitem{SemanticKITTI}
J.~Behley, M.~Garbade, A.~Milioto, J.~Quenzel, S.~Behnke, C.~Stachniss, and
  J.~Gall, ``{SemanticKITTI:} {A} dataset for semantic scene understanding of
  lidar sequences,'' in \emph{{IEEE/CVF} International Conference on Computer
  Vision}.\hskip 1em plus 0.5em minus 0.4em\relax {IEEE}, 2019, pp. 9296--9306.

\bibitem{long2015fully}
J.~Long, E.~Shelhamer, and T.~Darrell, ``Fully convolutional networks for
  semantic segmentation,'' in \emph{Proceedings of the IEEE conference on
  computer vision and pattern recognition}, 2015, pp. 3431--3440.

\bibitem{wang2020residual}
C.~Wang, W.~Pedrycz, Z.~Li, and M.~Zhou, ``Residual-driven fuzzy c-means
  clustering for image segmentation,'' \emph{IEEE/CAA Journal of Automatica
  Sinica}, vol.~8, no.~4, pp. 876--889, 2020.

\bibitem{wang2020g}
C.~Wang, W.~Pedrycz, Z.~Li, M.~Zhou, and S.~S. Ge, ``G-image segmentation:
  Similarity-preserving fuzzy c-means with spatial information constraint in
  wavelet space,'' \emph{IEEE Transactions on Fuzzy Systems}, vol.~29, no.~12,
  pp. 3887--3898, 2020.

\bibitem{wang2019wavelet}
C.~Wang, W.~Pedrycz, J.~Yang, M.~Zhou, and Z.~Li, ``Wavelet frame-based fuzzy
  c-means clustering for segmenting images on graphs,'' \emph{IEEE transactions
  on cybernetics}, vol.~50, no.~9, pp. 3938--3949, 2019.

\bibitem{shaban2017one}
A.~Shaban, S.~Bansal, Z.~Liu, I.~Essa, and B.~Boots, ``One-shot learning for
  semantic segmentation,'' in \emph{British Machine Vision Conference}.\hskip
  1em plus 0.5em minus 0.4em\relax {BMVA} Press, 2017.

\bibitem{Rakelly2018Conditional}
K.~Rakelly, E.~Shelhamer, T.~Darrell, A.~A. Efros, and S.~Levine, ``Conditional
  networks for few-shot semantic segmentation,'' in \emph{International
  Conference on Learning Representations(workshop)}.\hskip 1em plus 0.5em minus
  0.4em\relax OpenReview.net, 2018.

\bibitem{ZhangLLYS19}
C.~Zhang, G.~Lin, F.~Liu, R.~Yao, and C.~Shen, ``{CANet:} class-agnostic
  segmentation networks with iterative refinement and attentive few-shot
  learning,'' in \emph{{IEEE/CVF} Conference on Computer Vision and Pattern
  Recognition}.\hskip 1em plus 0.5em minus 0.4em\relax {IEEE}, 2019, pp.
  5217--5226.

\bibitem{SiamOJ19}
M.~Siam, B.~N. Oreshkin, and M.~J{\"{a}}gersand, ``{AMP:} adaptive masked
  proxies for few-shot segmentation,'' in \emph{{IEEE/CVF} International
  Conference on Computer Vision}.\hskip 1em plus 0.5em minus 0.4em\relax
  {IEEE}, 2019, pp. 5248--5257.

\bibitem{WangLZZF19}
K.~Wang, J.~H. Liew, Y.~Zou, D.~Zhou, and J.~Feng, ``{PANet:} few-shot image
  semantic segmentation with prototype alignment,'' in \emph{{IEEE/CVF}
  International Conference on Computer Vision}.\hskip 1em plus 0.5em minus
  0.4em\relax {IEEE}, 2019, pp. 9196--9205.

\bibitem{WangY0ZS021}
H.~Wang, Y.~Yang, X.~Cao, X.~Zhen, C.~Snoek, and L.~Shao, ``Variational
  prototype inference for few-shot semantic segmentation,'' in \emph{{IEEE}
  Winter Conference on Applications of Computer Vision}.\hskip 1em plus 0.5em
  minus 0.4em\relax {IEEE}, 2021, pp. 525--534.

\bibitem{YangLLJY20}
B.~Yang, C.~Liu, B.~Li, J.~Jiao, and Q.~Ye, ``Prototype mixture models for
  few-shot semantic segmentation,'' in \emph{European Conference on Computer
  Vision}, A.~Vedaldi, H.~Bischof, T.~Brox, and J.~Frahm, Eds., vol.
  12353.\hskip 1em plus 0.5em minus 0.4em\relax Springer, 2020, pp. 763--778.

\bibitem{ZhangXQ21}
B.~Zhang, J.~Xiao, and T.~Qin, ``Self-guided and cross-guided learning for
  few-shot segmentation,'' in \emph{{IEEE/CVF} Conference on Computer Vision
  and Pattern Recognition}.\hskip 1em plus 0.5em minus 0.4em\relax {IEEE},
  2021, pp. 8312--8321.

\bibitem{YanCXWLL19}
X.~Yan, Z.~Chen, A.~Xu, X.~Wang, X.~Liang, and L.~Lin, ``Meta {R-CNN:} towards
  general solver for instance-level low-shot learning,'' in \emph{{IEEE/CVF}
  International Conference on Computer Vision}.\hskip 1em plus 0.5em minus
  0.4em\relax {IEEE}, 2019, pp. 9576--9585.

\bibitem{KhandelwalGS21}
S.~Khandelwal, R.~Goyal, and L.~Sigal, ``{UniT:} unified knowledge transfer for
  any-shot object detection and segmentation,'' in \emph{{IEEE/CVF} Conference
  on Computer Vision and Pattern Recognition}.\hskip 1em plus 0.5em minus
  0.4em\relax {IEEE}, 2021, pp. 5951--5961.

\bibitem{TaveraCMC22}
A.~Tavera, F.~Cermelli, C.~Masone, and B.~Caputo, ``Pixel-by-pixel cross-domain
  alignment for few-shot semantic segmentation,'' in \emph{{IEEE/CVF} Winter
  Conference on Applications of Computer Vision}.\hskip 1em plus 0.5em minus
  0.4em\relax {IEEE}, 2022, pp. 1959--1968.

\bibitem{ZhaoFewCVPR2021}
N.~Zhao, T.~Chua, and G.~H. Lee, ``Few-shot 3d point cloud semantic
  segmentation,'' in \emph{{IEEE/CVF} Conference on Computer Vision and Pattern
  Recognition}.\hskip 1em plus 0.5em minus 0.4em\relax {IEEE}, 2021, pp.
  8873--8882.

\bibitem{Lai2022TacklingBA}
L.~Lai, J.~Chen, C.~Zhang, Z.~Zhang, G.~Lin, and Q.~Wu, ``Tackling background
  ambiguities in multi-class few-shot point cloud semantic segmentation,''
  \emph{Knowledge-Based Systems}, 2022.

\bibitem{Chen2020CompositionalPN}
X.~Chen, C.~Zhang, G.~Lin, and J.~Han, ``Compositional prototype network with
  multi-view comparision for few-shot point cloud semantic segmentation,''
  \emph{ArXiv}, vol. abs/2012.14255, 2020.

\bibitem{CorralSoto2022HYLDAEH}
E.~R. Corral-Soto, M.~Rochan, Y.~He, S.~Aich, Y.~Liu, and B.~Liu, ``{HYLDA:}
  end-to-end hybrid learning domain adaptation for lidar semantic
  segmentation,'' \emph{ArXiv}, vol. abs/2201.05585, 2022.

\bibitem{SalsaNext}
T.~Cortinhal, G.~Tzelepis, and E.~E. Aksoy, ``{SalsaNext:} fast,
  uncertainty-aware semantic segmentation of lidar point clouds,'' in
  \emph{International Symposium on Visual Computing}, G.~Bebis, Z.~Yin, E.~Kim,
  J.~Bender, K.~Subr, B.~C. Kwon, J.~Zhao, D.~Kalkofen, and G.~Baciu, Eds.,
  vol. 12510.\hskip 1em plus 0.5em minus 0.4em\relax Springer, 2020, pp.
  207--222.

\bibitem{BermanTB18}
M.~Berman, A.~R. Triki, and M.~B. Blaschko, ``The lov{\'{a}}sz-softmax loss:
  {A} tractable surrogate for the optimization of the intersection-over-union
  measure in neural networks,'' in \emph{{IEEE/CVF} Conference on Computer
  Vision and Pattern Recognition}.\hskip 1em plus 0.5em minus 0.4em\relax
  {IEEE}, pp. 4413--4421.

\bibitem{MiBCVPR2020}
F.~Cermelli, M.~Mancini, S.~R. Bul{\`{o}}, E.~Ricci, and B.~Caputo, ``Modeling
  the background for incremental learning in semantic segmentation,'' in
  \emph{{IEEE/CVF} Conference on Computer Vision and Pattern
  Recognition}.\hskip 1em plus 0.5em minus 0.4em\relax {IEEE}, 2020, pp.
  9230--9239.

\bibitem{TungM19}
F.~Tung and G.~Mori, ``Similarity-preserving knowledge distillation,'' in
  \emph{{IEEE/CVF} International Conference on Computer Vision}.\hskip 1em plus
  0.5em minus 0.4em\relax {IEEE}, 2019, pp. 1365--1374.

\bibitem{Li2018LearningWF}
Z.~Li and D.~Hoiem, ``Learning without forgetting,'' \emph{IEEE Transactions on
  Pattern Analysis and Machine Intelligence}, vol.~40, pp. 2935--2947, 2018.

\end{thebibliography}
\end{document}